# Covariate-distance Weighted Regression (CWR): A Case Study for Estimation of House Prices

**Hone-Jay Chu[1, *]    Po-Hung Chen[1]    Sheng-Mao Chang[2]    Muhammad Zeeshan Ali[1]**

**Sumriti Ranjan Patra[1]**

[1] Department of Geomatics, National Cheng Kung University, No. 1, University Road, Tainan City 701, Taiwan
[2] Department of Statistics, National Taipei University, No.151, University Road, New Taipei City 237, Taiwan

<u>Corresponding Author</u>[*]
Prof. Hone-Jay Chu, Associate Professor
Department of Geomatics
National Cheng Kung University
No. 1, University Road, Tainan City 701, Taiwan
honejaychu@geomatics.ncku.edu.tw

## Abstract:

Geographically weighted regression (GWR) is a popular tool for modeling spatial heterogeneity in a regression model. However, the current weighting function used in GWR only considers the geographical distance, while the attribute similarity is totally ignored. In this study, we proposed a covariate weighting function that combines the geographical distance and attribute distance. The covariate-distance weighted regression (CWR) is the extension of GWR including geographical distance and attribute distance. House prices are affected by numerous factors, such as house age, floor area, and land use. Prediction model is used to help understand the characteristics of regional house prices. The CWR was used to understand the relationship between the house price and controlling factors. The CWR can consider the geological and attribute distances, and produce accurate estimates of house price that preserve the weight matrix for geological and attribute distance functions. Results show that the house attributes/conditions and the characteristics of the house, such as floor area and house age, might affect the house price. After factor selection, in which only house age and floor area of a building are considered, the RMSE of the CWR model can be improved by 2.9%-26.3% for skyscrapers when compared to the GWR. CWR can effectively reduce estimation errors from traditional spatial regression models and provide novel and feasible models for spatial estimation.

# 1. Introduction

Various methods have been applied for house price estimation, such as statistical regression ([*Lu et al.*, 2011; *Y Wang et al.*, 2017; *Wu et al.*, 2018]), neural networks ([*Limsombunchai*]), machine learning (ML)[*Park and Bae*, 2015], such as random forest [*Chen et al.*, 2014], support vector machine [*Bui et al.*, 2015], and least squares boosting (LSBoost; [*Bühlmann and Hothorn*, 2007]). On the basis of the relation of house price–house factors from the neighboring houses, house price can be estimated based on known house factors. Hedonic price models assume that the price of a product reflects embodied characteristics valued by some implicit or shadow prices ([*Limsombunchai*]; [*Owusu-Ansah*, 2011]; [*Randeniya et al.*, 2017]). Hedonic models are most commonly estimated using a regression model. Regression-based analysis is a conventional approach for prediction ( [*Lu et al.*, 2011] ; [*X Wang et al.*, 2014]; [*Wu et al.*, 2018] ). Linear regression is a simple and interpretable tool for house price prediction. Given a linear model, one may determine how having an additional bathroom affects the house price. However, linear regression blurs the effect arising from local environment; for example, having an additional bathroom has various effects in different locations. By contrast, geographically weighted regression (GWR; [*Fotheringham et al.*, 2003]; GTWR [*Huang et al.*, 2010]) consider the geographical weight with Euclidean distances among observations. The GWR can capture the effects of spatially heterogeneous processes and explore nonstationary relations in data-generating processes by allowing regression coefficients to vary spatially. The GWR allows geographical weighting (e.g., heterogeneous or cluster effects on estimation), whereas linear regression does not. However, the current weighting function used in GWR only considers the geographical distances, while the attribute similarity is totally ignored. In this study, we proposed a covariate weighting function that combines the geographical distance and attribute distance. The spatial weights matrix is considered by distance and attribute with measures of each observation's neighborhood (Harris et al., 2013). The proposed covariate-distance weighted regression (CWR) is the extension of GWR via enrichment of the dimension of distance, i.e. geographical distance and attribute distance. Geographical location and house attributes affect house prices ([*Schiele*, 2008]). We proposed a new weighting function that combines the geographical and attribute distances. The spatial-attribute weighting function would be better than that of the GWR model with the spatial weighting function [*Moore and Myers*, 2010; *Shi et al.*, 2006]. The weight kernel of CWR is the function of multiple distances, which hybridizes geographical distance and the attribute distance in the houses.

The CWR is developed to estimate house prices, and mainly focus on the combination of geographical and attribute distances on weighting functions. The CWR is used to model the spatial relationship between the house price and factors, and find the optimal rate of distance weights. The model performance of CWR model is checked when compared with LR, GWR, and ML. The rest of this paper is organized as follows. Details of the CWR is addressed next, followed by our results and conclusion.

# 2. Materials and study area

To promote transparency of transaction data in the real estate market and prevent the harmful bidding up of real estate prices, the Taiwanese government has implemented a "real price registration" system from open data. Registration of real price contains attributes (e.g., address, building area, building pattern, trading time, construction time, construction type, and total price), which are used as factors. In addition, the transaction data used in this study are under the conditions of various districts and house types in Taichung City during 12 seasons from 2014 to 2016. Taichung, the study area, is located along the west coast of Taiwan. Taichung has a pleasant climate and environment. It has a population of approximately 2.81 million people. Taichung is the second largest metropolitan area and most populous city in Taiwan. In this study, four districts, namely, North, South, Dali, and Beitung Districts, are considered (Figure 1).

The house prices used in this study are the open data products in Taiwan from the Ministry of the Interior (https://lvr.land.moi.gov.tw/homePage.action). Moreover, this study not only considers the property of the house but also the degree of influence of points of interest (POI)/external factors to the price. We consider the Euclidean distance between the house and POI location (e.g., museum, library, hotel, convenience store, and gas station). Regarding land use, we also consider the type of land use where the house is located and used dummy variables for various categories. Data include 431, 764, 262, and 1035 skyscrapers; 662, 178, 170, and 333 apartments; and 497, 151, 417, and 694 for luxury condos in North, South, Dali, and Beitung Districts, respectively. The datasets are independently set as the training and testing data, which contain 80% and 20%, respectively. After the preprocessing, the data are then applied to the four different models (i.e., Least Squares: LS, GWR, CWR, and ML models). Dependent data include floor area, house age, distances to museum/ library/ hotel/ convenience store/ train station/ school/ gas station/ temple/ police station/ restaurant/ parking lot, number of rooms, number of bathrooms, and number of living rooms for house estimation.

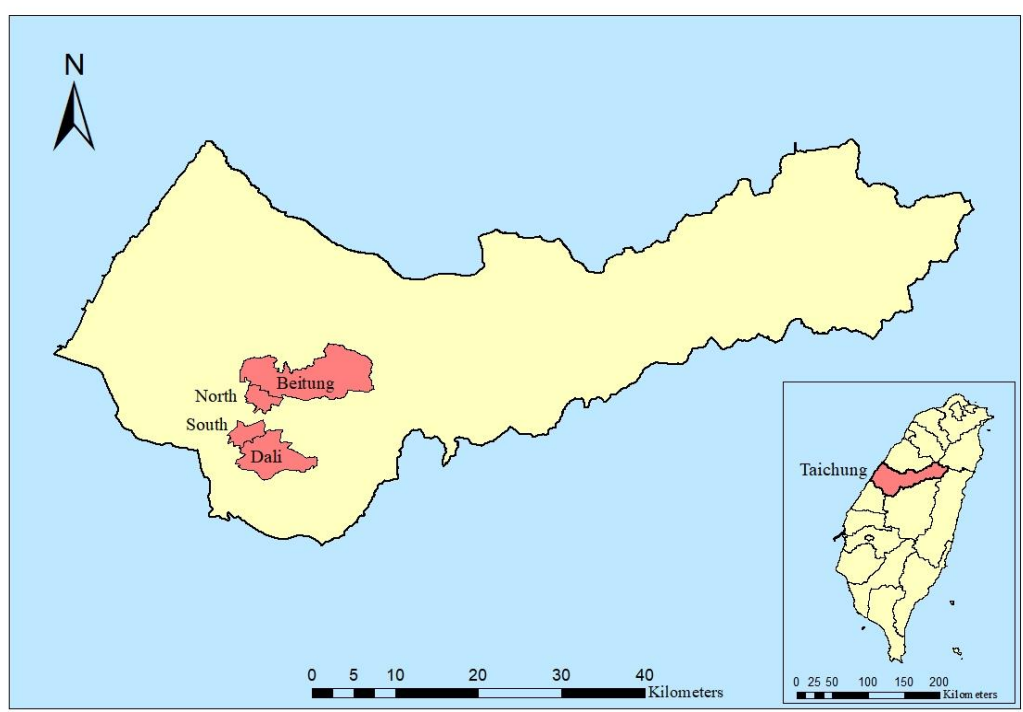

**Figure 1**. Location of the study area for the four Districts

## 3. Methods

Figure 2 shows the flowchart of the model. Firstly, factor selection is used for the factor identification of house price estimation. The house price estimations in three house types (i.e., skyscrapers, apartments, and luxury condos) for four districts are applied using CWR. In the algorithm, the best combination of spatial distance and attribute distance is determined. Finally, the LS, GWR, and ML models are used to for comparison. Model comparison is conducted through the predicted values produced by each model from root mean squared error (RMSE).

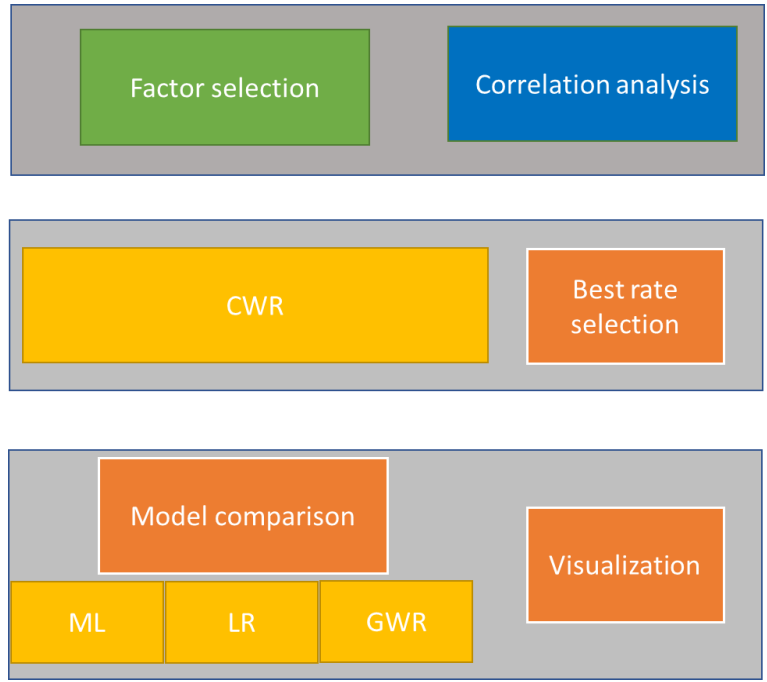

**Figure 2.** Flowchart of this study 1. Correlation analysis 2. CWR model development 3. Model comparison and visualization

### 3.1. Factor selection

The predictor importance function is used to recognize the importance factors. The predictor importance is measured based on the amount of improved prediction performance, that is, mean squared error (MSE) change when splitting data points on that predictor ([*Su et al.*, 2019]). The more a predictor is used with these trees, the higher its importance will be. After the predictor importance function computes estimates of predictor importance for tree by summing the changes in the MSE, the important factors can be sorted.

### 3.2. GWR and CWR model

Linear Regression is a model that attempts to model a linear relationship between dependent variables, $Y_i$ and independent variables, $X_i$. $\beta_0$ is the intercept, and $\beta_b, b = 1, \ldots, B$. is the slope of the model.

$$Y_i = \beta_0 + \sum_{b=1}^{B} \beta_b x_{bi} \qquad (1)$$

The GWR model is defined as follows. Suppose that $n$ observations are found in the dataset. Each observation consists of a response variable $Y_i$ and $B$-dimensional covariates $X_i = (x_{1i}, \ldots, x_{Bi})^T$, $i = 1, \ldots, n$. GWR requires the 2-D coordinates of each observation as geographical information. Let $(u_i, v_i)$ be the coordinate of the $i$th observation. Given a particular coordinate, say $(u_0, v_0)$, the GWR assumes

$$Y_i = \beta_0(u_0, v_0) + \sum_{b=1}^{B} \beta_b(u_0, v_0) x_{bi} \qquad (2)$$

Regression coefficients, $\beta_0(u_0, v_0), \ldots, \beta_b(u_0, v_0)$, are estimated by weighted least-square method, where the weights are evaluated by the distance between coordinates of the observed data $(u_i, v_i)$ and the specified coordinate $(u_0, v_0)$. Popular distances are Euclidian and Gaussian distances, among others. In GWR, the estimated coefficient $\hat{\beta}_k(u_i, v_i)$ at each observation $i$ is expressed as follows:

$$\hat{\beta}_k(u_i, v_i) = [X^T W(u_i, v_i) X]^{-1} X^T W(u_i, v_i) Y \qquad (3)$$

where $W(u_i, v_i)$ is a weight matrix based on the Euclidean and Gaussian distance decay-based functions in the spatial feature domains. Gaussian distance decay-based function is one of the typical function of spatial autocorrelation[*Huang et al.*, 2010]). The geographical distance is quoted in meters. The elements in a weight matrix $W_{ij}$ is the weighted element between observations $i$ and $j$. The weight element can be computed as

$$W_{ij} = exp\left(-\frac{d_{ij}^2}{h^2}\right) \qquad (4)$$

where $h$ is a nonnegative parameter known as bandwidth. In the GWR, $d_{ij}$ is Gaussian distance decay-based function i.e. $d_{ij}^g$ based on geographical distance only. The bandwidth can be determined by several criteria, such as cross validation ([*Brunsdon et al.*, 1996]).

The CWR formation is also defined as the Eq (2), and the estimated coefficient $\hat{\beta}_k(u_i, v_i)$ can be as the Eq (3). Particularly, the weight matrix is based on the Euclidean and Gaussian distance decay-based functions in the space–attribute domains. The CWR simultaneously considers the geographical and attribute distances, as shown in Figure 3. Geographical distance is based on longitude and latitude information, and the attribute distance of the two samples is based on attributes, such as the house age and floor area. In weight element (Eq (4)), the trade-off distances include the geographical and attribute distances. That is:

$$d_{ij} = r \times d_{ij}^g + (1 - r) \times d_{ij}^a \qquad (5)$$

where $d_{ij}^g$ is the geographical distance between houses $i$ and $j$, and $d_{ij}^a$ is the distance in attribute between houses $i$ and $j$. Rate $r$ is the parameter for determining the component of the distance. The CWR model is the GWR if the ratio $r$ is equal to one. $r$ is the decision variable from the linear search that is decided by the criteria for minimum of RMSE based on training data. The best rate $r$ between geographical and attribute distances is acquired in this study.

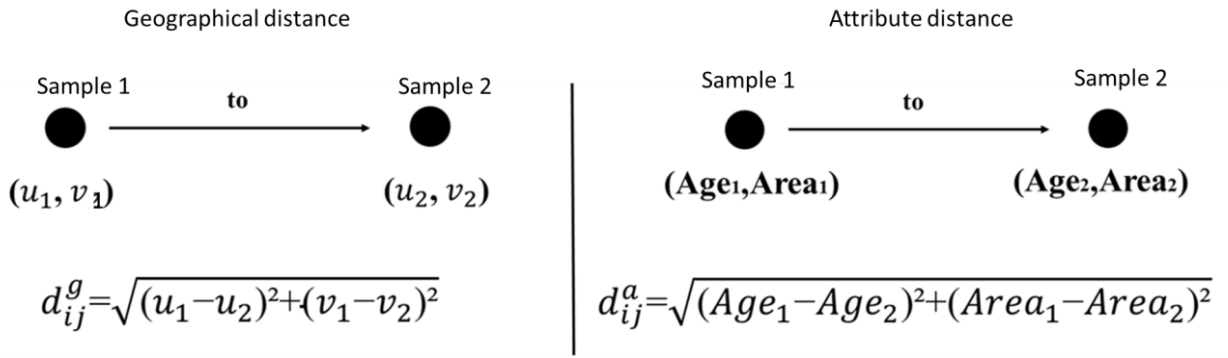

**Figure 3.** Weighting evaluations from geographical and attribute distances

Given a particular location $(u_0, v_0)$ with house attribute $X_0$, we aim to estimate the unknown house price $(\hat{Y}_0)$ in the neighbor area based on the K-nearest neighbor (KNN) strategy. K=3 is used in this case study.

### 3.3. ML method- LSBoost

The LSBoost method, one of the most popular ML methods, is used to address the regression problem ([*Su et al.*, 2019]). LSBoost fits regression ensembles for least-square problems. LSBoost applies the least squares as the loss criteria and can thus fit regression ensembles well to minimize the MSEs. The model utilizes boosting in ensemble learning with decision trees. This ML method is used for comparison and validation of models in terms of predictive performance of house prices.

# 4. Results

## 4.1. Factor selection and correlation analysis

Figure 4 shows the ranking of important factors using decision trees. Important factor is highly related to the house price. The results show that the floor area and house age are critical factors in the prediction of house prices. We use these factors as the major independent variables in further usages. Distances to the POI such as distances to museum, convenience store and train station are unsuitable for the identification of house price. However, POI is not critical in the district. For example, in Taiwan, convenience stores are ubiquitous in the urban area. Reports say that Taiwan has the highest density of convenience stores in the world ([*Chang and Meyerhoefer*, 2018]). Convenience stores are a modern part of life in Taiwan. Accessibility of the neighborhood is nearly equal in the districts in the urban area of Taiwan. Therefore, distances to the POI is not an important factor in this case. However, house price is highly affected by proximity to the museum compared with the convenience store because museums are few and unique in the area. The result matches with the previous study, which indicates that proper infrastructure and facilities in the housing area can increase housing prices ([*Kamal et al.*, 2016]). House price is highly affected by physical characteristics of a property, such as floor area and house age. However, number of rooms, bathroom and living rooms are not the important factors for house price prediction. Many studies have mentioned that the age of buildings is negatively correlated with the value of residential properties, whereas the size of buildings is positively correlated with the prices ( [*Ferlan et al.*, 2017]; [*Kamal et al.*, 2016]). We identify the attributes e.g. floor area and house age that correlate with housing prices as the most significant attributes ( [*Randeniya et al.*, 2017]).

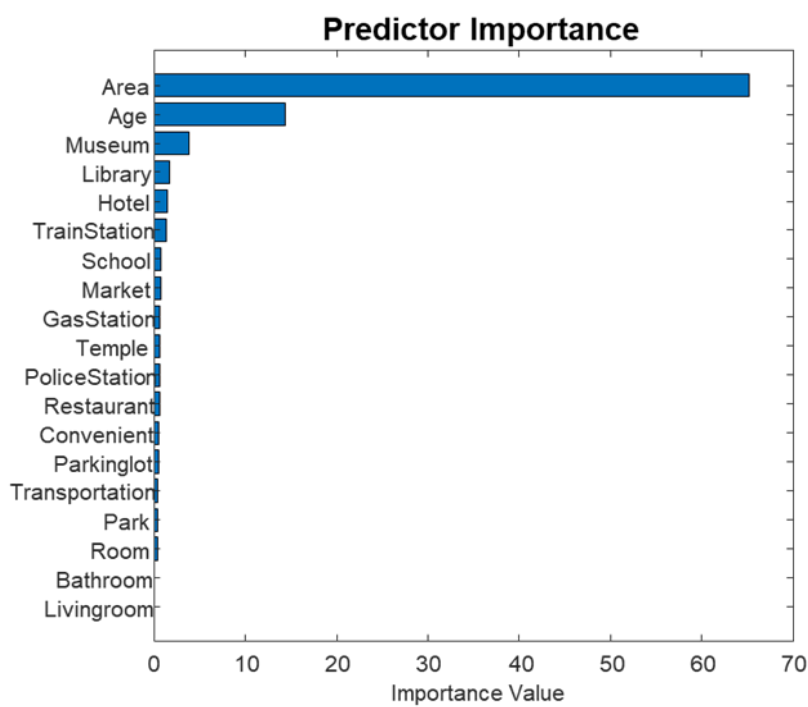

**Figure 4.** Ranking of important factors i.e. floor area, house age, distances to museum, library, hotel, convenience store, train station, school, marker, gas station, temple, police station, restaurant, parking lot, park, number of rooms, number of bathrooms, and number of living rooms for house estimation

## 4.2. CWR performance

Table 1 shows the model RMSE of each case in this study. In the experiments, the actual and estimated house prices are utilized as evaluation indicators of model performance. The values in bold indicate the lowest value in the case. The result shows that the CWR model performs better than the other models in most cases. Result shows the RMSE of the LR, GWR, CWR, and ML, respectively. In the comparison, the performance of the LR model is usually good in the South District, whereas the estimation error of skyscrapers is large in LR. The fitted models can also be easily interpreted ( [*Owusu-Ansah*, 2011]). Overall, the result that the estimation of CWR is better than LR and GWR methods seems reasonable. In comparison with LR, CWR improve about 70% mostly for skyscrapers in Beitung. In comparison with GWR, CWR improves 7.9% in luxury condo in Dali District. The improvement of the CWR reaches 2.9%-26.3% in skyscrapers, especially in skyscrapers in Beitung District. In sum, house price estimation is more practical and effective via CWR compared with GWR, considering the weight function of attribute and geological distances. The results indicate that the proposed CWR model has good forecasting performance for house price estimation.

**Table 1.** Estimation RMSE of each case in this study for skyscrapers, apartments, and luxury condos in the four districts using ML, LR, GWR and CWR. Bold indicates the lowest value in the case (unit: 10k NTD)

| | | North | South | Dali | Beitung |
|---|---|---|---|---|---|
| Skyscrapers | ML | 115.0 | 98.1 | 93.3 | 132.0 |
| | LR | 128.4 | 97.7 | 116.7 | 211.0 |
| | GWR | 100.0 | 82.9 | 82.7 | 153.5 |
| | CWR | **93.0** | **80.6** | **76.7** | **121.5** |
| Apartments | ML | 121.0 | 110.2 | 72.1 | 90.9 |
| | LR | 109.6 | 75.2 | 68.4 | 83.5 |
| | GWR | 109.4 | **75.2** | 68.4 | **83.5** |
| | CWR | **109.3** | 75.3 | **68.2** | 83.5 |
| Luxury Condos | ML | 106.1 | 81.2 | **73.5** | 96.6 |
| | LR | 100.2 | 73.4 | 88.4 | 94.7 |
| | GWR | 100.1 | **73.4** | 82.8 | 83.0 |
| | CWR | **99.8** | 73.4 | 76.7 | **82.8** |

## 4.2.1 CWR, GWR and LR

Figure 5 shows the overall comparison of CWR with the GWR and LR regression methods. The overall performance of CWR proves that the combination of attribute and geographical distances for the weight evaluation can effectively improve the shortcomings of GWR and LR model, especially for skyscrapers. House price of skyscrapers is highly related to the attribute distance.

We compare three house types in the four districts of study area. The results indicate that the CWR model is feasible and effective for estimating house prices. LR can be used when the attributes are highly related to house price (e.g., in South District). Moreover, the CWR and GWR models improve the LR and estimates the spatially varying relationship between the factors and house prices. The CWR combines the similarity of house attributes and spatially varying function to estimate reliable house prices. In comparison with GWR, the skyscrapers show the most improvement using CWR because its attribute is useful, especially in Beitung. This comparison result is caused by the floor area and age of skyscrapers effectively affecting the house prices. Thus, the CWR model performance is highly reliable. The CWR model of the skyscrapers is the most positive performance compared with ML. The attribute-based distance weight is more critical than that based on geographical distance. House price can be estimated accurately based on the attributes of houses. The weighting evaluation for hybrid distance set as the function of additional attribute distances is effective in the CWR model.

Estimation of house prices in Beitung shows the most improvement because the house price in the area shows serious heterogeneity such as old and new buildings. Increased functional flexibility pays off in terms of substantially reliable predictions of housing price if the geography is not an important factor or geography is in a multicentric and heterogenous city ([*Osland et al.*, 2007]).

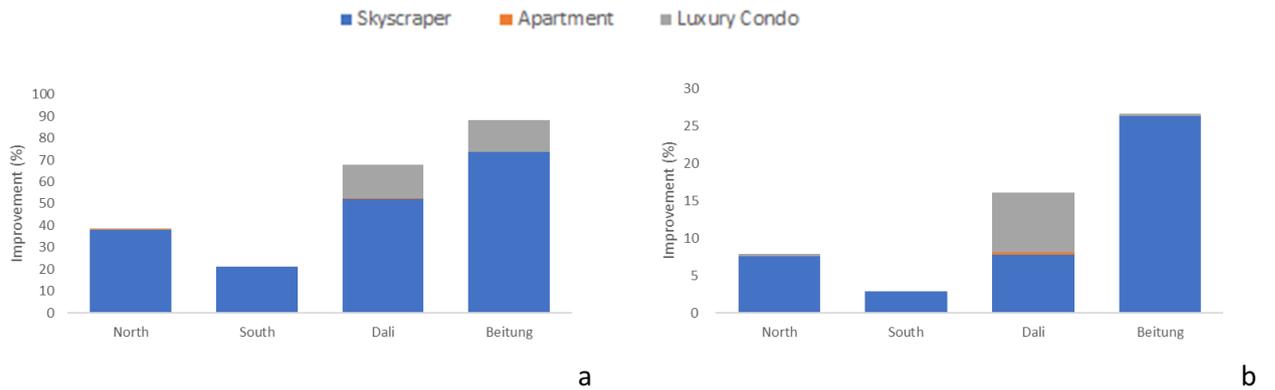

**Figure 5.** CWR improvement when compared with (a) LR, (b) GWR with various building types in the four district areas

### 4.2.2 CWR and ML

Figure 6 shows the comparison of CWR and ML model. The CWR model contains better performance about 14% compared with ML (6~23%), without for apartment in South District. ML overestimates the house prices of apartments in South District (46%). The negative performance of CWR is shown in only Dali, which is less than 5% compared with the ML for apartment.

Figure 7 shows the estimated maps house price for three house types in North District. Figure 8 shows the CWR and ML model residuals in skyscrapers, apartments, and luxury condos in North District. Spatial pattern of residuals shows random patterns. Spatial patterns of residuals from CWR and ML are consistent. Thus, CWR and ML methods are available because the residuals are independent of the methods. Large RMSEs occur at only few locations of estimation. The performance of CWR will be better or similar to the ML model. The CWR will be more powerful and the CWR will reach the ML level when considering the attribute distance and geographical distance weights. In the future, other machine learning approaches, e.g. XgBoost, Catboost, adaBoost, will have strong implications on the model performance.

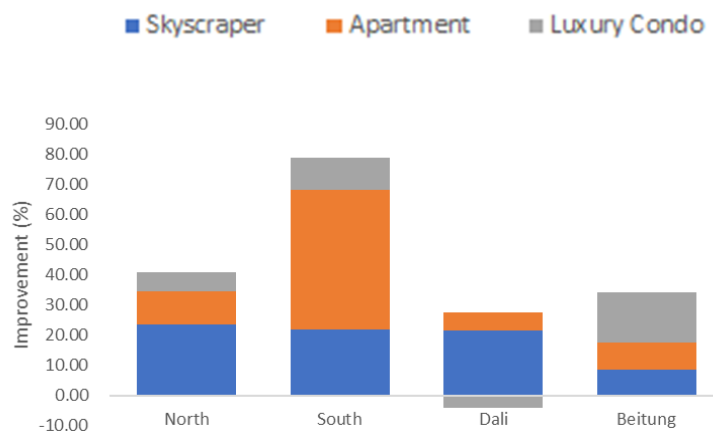

**Figure 6.** CWR improvement when compared with ML with various building types in the four district areas

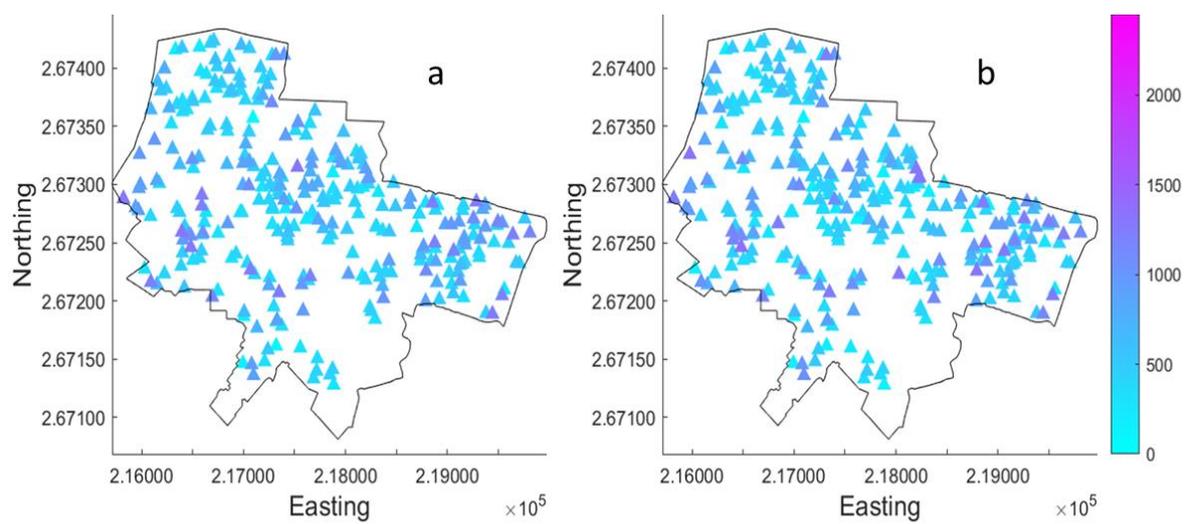

**Figure 7.** House price estimated maps in North District using (a)CWR and (b)ML (unit: 10k NTD)

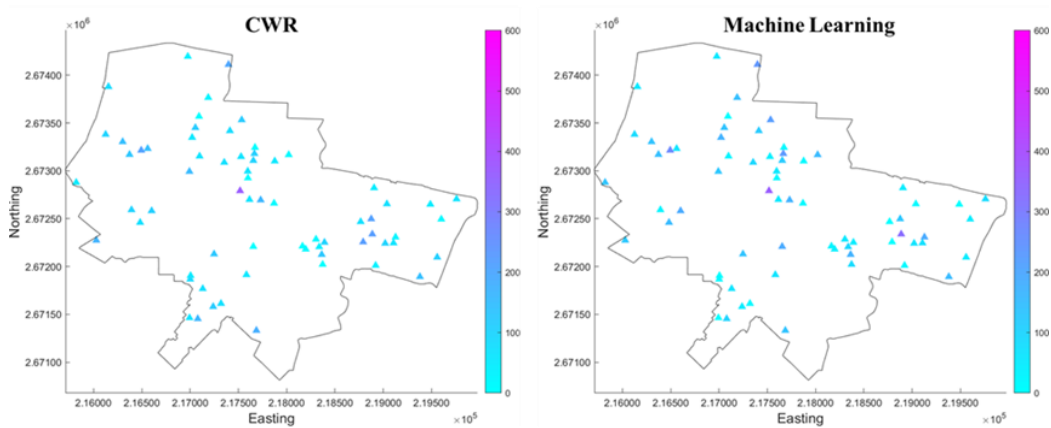

(a)

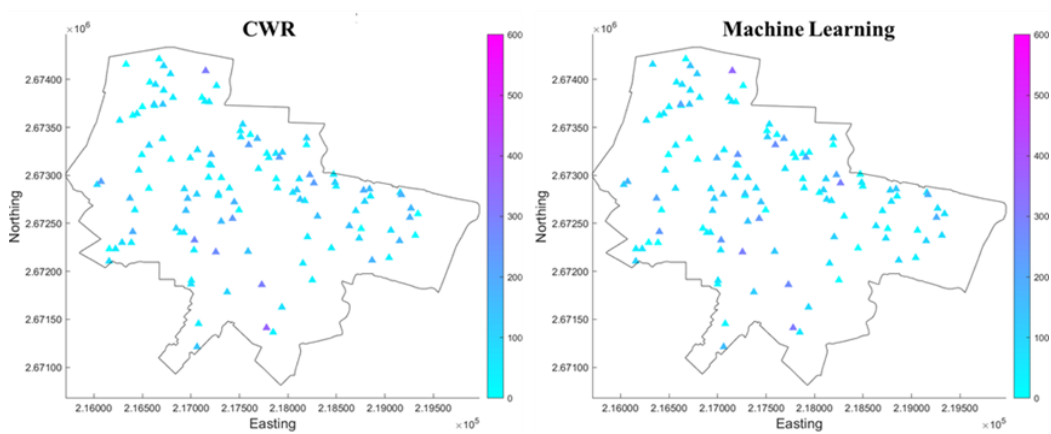

(b)

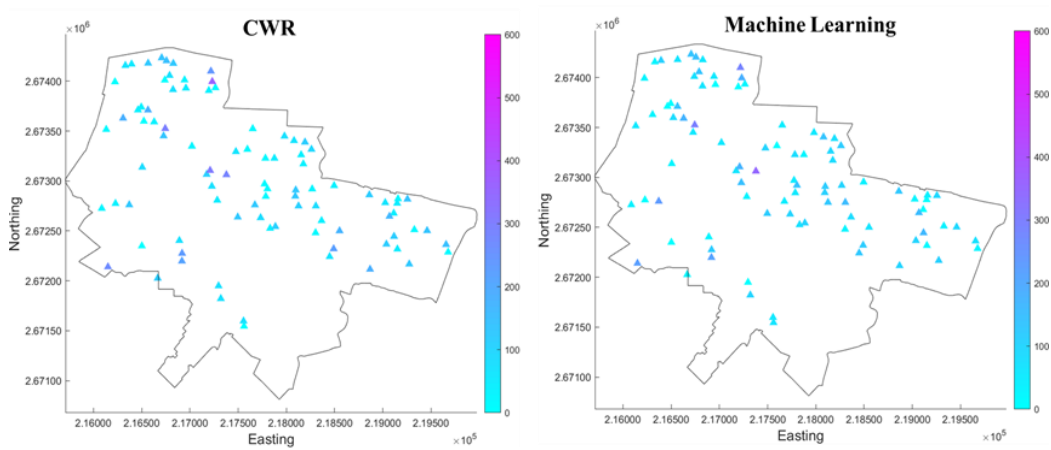

(c)

**Figure 8.** Model residual maps using CWR and ML for (a) skyscrapers, (b) apartments, and (c) luxury condos in North District (unit: 10k NTD)

### 4.3. Sensitivity analysis

Additional calibration of the model is not required, which is different from typical ML models that usually require an optimization modeling component for calibration purposes. The fundamental benefit of the CWR model is its ability to automatically estimate parameter values that produce the best fit. Based on the optimal weight, the CWR estimation error is the lowest in the most cases. Table 2 shows the best ratio $r$ between geographical and attribute distances. If the ratio $r$ is close to 1, the CWR is the original GWR. For example, the estimation error is the minimum value at apartment and luxury condos in District South, apartments in District Beitung and District South when the weight is equal to 1 in the CWR. However, for skyscrapers in District North, District South and District Dali, the CWR considers the small $r$ for highlighting on the attribute distance in Eq (5). However, limitations include data uncertainty (e.g., address missing and geographic masking). Future study will consider the uncertainty analysis of model. Furthermore, a temporal weighted approach will consider estimation of temporal and nonstationary house prices[*Huang et al.*, 2010]. In addition, complexity of real world ensures that the scope of possible distance metrics is far larger than the traditional Euclidean distance. Road network distance or travel distance can be considered in the future [*Cao et al.*, 2019; *Lu et al.*, 2014].

**Table 2. Optimal ratio coefficient $r$ in the CWR**

|                | North | South | Dali | Beitung |
|----------------|-------|-------|------|---------|
| Skyscrapers    | .16   | .25   | .23  | .96     |
| Apartments     | .13   | 1     | .99  | 1       |
| Luxury Condos  | .98   | 1     | .16  | .94     |

# 5. Discussion

Housing market reflects the current economic situation and social sentiments in the country ([Ng and Deisenroth, 2015]). Prediction of house prices is crucial for buyers and owners, especially in large cities. At present, people obtain information from housing market to determine the price in a regional area as a reference for real estate sales. Such limited information may cause bias in the transaction. Moreover, providing effective predictions through the actual data of price registration is useful for the estimation of housing prices. Predicting house values can help in understanding the characteristics of the area's housing market. House price prediction is challenging because house price is affected by many global factors, such as government policy, and economic growth ([Phang and Wong, 1997]). Surrounding POI or regional factors, such as school, traffic and natural scenery ([Wu et al., 2018]) affect house price estimating. In addition, local or individual factors such as land use may also affect the price significantly. These factors may include the size and age of houses ([Xu et al., 2018]); house type; number of bedrooms, bathrooms, and garages; amenities around the house; and geographical location ([Limsombunchai]).

This study proves the availability of the weighting evaluation of additionally using attribute similarity. Geographical distance is generally critical because the first law of geography is that everything is related to everything else, but near objects are more related to distant ones ([*Tobler*, 1970]). However, house price similarity is not highly dependent on the geographical distances of houses. From the test case, house attribute is more critical rather than geographical distance. In the study area, a skyscraper is mixed with the old departments. Although two houses are geographically close, the contexts (attributes) within which they are not alike (Harris et al., 2013). Obviously, the house prices in the neighbors are not mainly correlated in the case. The weighting evaluation for hybrid distance is set as the function of attribute and geographical distances in the CWR model. This CWR model considers the trade-off between the geographical distance and attribute distance to estimate the house price for any cases. Therefore, the CWR is the extension of geographically weighted regression which only considers the geological distance. The use of geographical distance only for determining the weights of the GWR model may not be realistic [*Shi et al.*, 2006].

## 6. Conclusions

This study develops the CWR to estimate the relationship between the house price and its factors. The CWR can capture the effects of spatially heterogeneous processes; it can also explore spatial nonstationary in estimation processes by allowing regression spatially varying coefficients effectively and produce accurate estimates of house price to preserve the weight matrix for geological and attribute distance functions.

The results show that the conditions and the characteristic of the house, namely, house age and floor area, affect the house price. Additional house attributes provide information for the regression parameter weight generation. The weight matrix is based on Euclidean and Gaussian distance decay-based functions in the space–attribute domains. In comparison with the GWR, the results show that the model performance of the CWR is usually better than that of the GWR (2.9%-26.3% improvement). Given the house attributes for regression weight adjustment, most cases using the CWR has improved compared with the traditional GWR. Therefore, the proposed method in this study overcomes the shortcomings of the traditional GWR model and is easier to utilize in the real-world applications. This CWR method can effectively obtain satisfactory weight balance between geographical and attribute distances.

## Acknowledgment:

The authors would like to thank the GeoSmart Lab members and coworkers: Po-Hung Chen for coding; Muhammad Zeeshan Ali**,** Sumriti Ranjan Patra and Dr. S-M Chang for providing helps and suggestions for paper improvement. The study was supported by Ministry of Science and Technology (MOST), Taiwan (109-2621-M-006 -003 -).

## References:

1. Bühlmann, P., and T. Hothorn (2007), Boosting algorithms: Regularization, prediction and model fitting, *Statistical Science*, *22*(4), 477-505.
2. Brunsdon, C., A. S. Fotheringham, and M. E. Charlton (1996), Geographically weighted regression: a method for exploring spatial nonstationarity, *Geogr. Anal.*, *28*(4), 281-298.
3. Bui, D. T., B. Pradhan, I. Revhaug, D. B. Nguyen, H. V. Pham, and Q. N. Bui (2015), A novel hybrid evidential belief function-based fuzzy logic model in spatial prediction of rainfall-induced shallow landslides in the Lang Son city area (Vietnam), *Geomatics, Natural Hazards and Risk*, *6*(3), 243-271.

4.  Cao, K., M. Diao, and B. Wu (2019), A Big Data–Based Geographically Weighted Regression Model for Public Housing Prices: A Case Study in Singapore, *Annals of the American Association of Geographers*, *109*(1), 173-186.

5.  Chang, H.-H., and C. D. Meyerhoefer (2018), Inter-brand Competition in the Convenience Store Industry, Store Accessibility and Healthcare Utilization*Rep.*, National Bureau of Economic Research.

6.  Chen, W., X. Li, Y. Wang, G. Chen, and S. Liu (2014), Forested landslide detection using LiDAR data and the random forest algorithm: A case study of the Three Gorges, China, *Remote sensing of environment*, *152*, 291-301.

7.  Ferlan, N., M. Bastic, and I. Psunder (2017), Influential Factors on the Market Value of Residential Properties, *Engineering Economics*, *28*(2), 135-144.

8.  Fotheringham, A. S., C. Brunsdon, and M. Charlton (2003), *Geographically weighted regression: the analysis of spatially varying relationships*, John Wiley & Sons.

9.  Harris, R., Dong, G., & Zhang, W. (2013). Using Contextualized G eographically W eighted R egression to Model the Spatial Heterogeneity of Land Prices in B eijing, C hina. *Transactions in GIS*, *17*(6), 901-919.

10. Huang, B., B. Wu, and M. Barry (2010), Geographically and temporally weighted regression for modeling spatio-temporal variation in house prices, *International Journal of Geographical Information Science*, *24*(3), 383-401.

11. Kamal, E. M., H. Hassan, and A. Osmadi (2016), Factors influencing the housing price: developers' perspective, *International Journal of Social, Behavioral, Educational, Economic, Business and Industrial Engineering*, *10*(5), 1603-1609.

12. Limsombunchai, V. House price prediction: hedonic price model vs. artificial neural network, 2004.

13. Lu, B., M. Charlton, and A. S. Fotheringhama (2011), Geographically weighted regression using a non-Euclidean distance metric with a study on London house price data, *Procedia Environmental Sciences*, *7*, 92-97.

14. Lu, B., M. Charlton, P. Harris, and A. S. Fotheringham (2014), Geographically weighted regression with a non-Euclidean distance metric: a case study using hedonic house price data, *International Journal of Geographical Information Science*, *28*(4), 660-681.

15. Moore, J. W., and J. Myers (2010), Using geographic-attribute weighted regression for CAMA modeling, *Journal of Property Tax Assessment & Administration*, *7*(3), 5-28.

16. Ng, A., and M. Deisenroth (2015), Machine learning for a London housing price prediction mobile application, in *Imperial College London*, edited.

17. Osland, L., I. Thorsen, and J. Gitlesen (2007), Housing price gradients in a region with one dominating center, *Journal of real estate research*, *29*(3), 321-346.

18. Owusu-Ansah, A. (2011), A review of hedonic pricing models in housing research, *Journal of International Real Estate and Construction Studies*, *1*(1), 19.

19. Park, B., and J. K. Bae (2015), Using machine learning algorithms for housing price prediction: The case of Fairfax County, Virginia housing data, *Expert Systems with Applications*, *42*(6), 2928-2934.

20. Phang, S.-Y., and W.-K. Wong (1997), Government policies and private housing prices in Singapore, *Urban studies*, *34*(11), 1819-1829.

21. Randeniya, T. D., G. Ranasinghe, and S. Amarawickrama (2017), A model to estimate the implicit values of housing attributes by applying the hedonic pricing method, *International Journal of Built Environment and Sustainability*, *4*(2).

22. Schiele, H. (2008), Location, location: the geography of industry clusters, *Journal of business strategy*, *29*(3), 29-36.

23. Shi, H., L. Zhang, and J. Liu (2006), A new spatial-attribute weighting function for geographically weighted regression, *Canadian journal of forest research*, *36*(4), 996-1005.

24. Su, M., Z. Zhang, Y. Zhu, and D. Zha (2019), Data-Driven Natural Gas Spot Price Forecasting with Least Squares Regression Boosting Algorithm, *Energies*, *12*(6), 1094.

25. Tobler, W. R. (1970), A computer movie simulating urban growth in the Detroit region, *Economic geography*, *46*(sup1), 234-240.

26. Wang, X., J. Wen, Y. Zhang, and Y. Wang (2014), Real estate price forecasting based on SVM optimized by PSO, *Optik-International Journal for Light and Electron Optics*, *125*(3), 1439-1443.

27. Wang, Y., S. Wang, G. Li, H. Zhang, L. Jin, Y. Su, and K. Wu (2017), Identifying the determinants of housing prices in China using spatial regression and the geographical detector technique, *Applied Geography*, *79*, 26-36.

28. Wu, H., H. Jiao, Y. Yu, Z. Li, Z. Peng, L. Liu, and Z. Zeng (2018), Influence factors and regression model of urban housing prices based on internet open access data, *Sustainability*, *10*(5), 1676.

29. Xu, Y., Q. Zhang, S. Zheng, and G. Zhu (2018), House Age, Price and Rent: Implications from Land-Structure Decomposition, *The Journal of Real Estate Finance and Economics*, *56*(2), 303-324.